\documentclass[a4paper,fleqn]{cas-dc}

\usepackage[numbers]{natbib}
\usepackage[utf8]{inputenc}
\usepackage[T1]{fontenc}
\usepackage[tight,footnotesize]{subfigure}
\usepackage{graphicx}
\usepackage{url}
\usepackage{balance}
\usepackage{rotating}
\usepackage{xcolor}
\usepackage{hyperref}

\usepackage{cleveref}
\usepackage{amssymb}
\usepackage{amsmath}
\usepackage{array}
\usepackage{booktabs}
\usepackage{nicematrix}
\usepackage{mathtools}
\usepackage{array} 
\usepackage{varwidth} 
\usepackage{adjustbox}
\usepackage[ruled,vlined,linesnumbered]{algorithm2e}

\usepackage{color, colortbl}

\DeclarePairedDelimiter\ceil{\lceil}{\rceil}

\newcommand{\nosemic}{\renewcommand{\@endalgocfline}{\relax}}
\newcommand{\dosemic}{\renewcommand{\@endalgocfline}{\algocf@endline}}
\let\oldnl\nl
\newcommand{\nonl}{\renewcommand{\nl}{\let\nl\oldnl}}

\def\tsc#1{\csdef{#1}{\textsc{\lowercase{#1}}\xspace}}
\tsc{WGM}
\tsc{QE}
\tsc{EP}
\tsc{PMS}
\tsc{BEC}
\tsc{DE}

\newcommand{\NAME}{ACSP-FL}
\newcommand{\COMPLETENAME}{\textbf{A}daptive \textbf{C}lient \textbf{S}election with \textbf{P}ersonalization for communication efficient  \textbf{F}ederated \textbf{L}earning}

\RequirePackage{xstring}
\RequirePackage{xparse}
\RequirePackage[]{acro}
\usepackage{xpatch}

\makeatletter
\@ifpackagelater{acro}{2019/02/17}{
    \@ifpackagelater{acro}{2020/04/29}{
        \NewDocumentCommand\acrodef{mO{#1}mG{}}{\DeclareAcronym{#1}{short={#2}, long={#3}, #4}}
    }{
        \NewDocumentCommand\acrodef{mO{#1}mG{}}{\DeclareAcronym{#1}{short={#2}, long={#3}, foreign-plural={}, #4}}
    }
}{
    \NewDocumentCommand\acrodef{mO{#1}mG{}}{\DeclareAcronym{#1}{short={#2}, long={#3}, #4}}
}
\makeatother
\AtEndDocument{\acuseall} 

\acrodef{MEC}{Mobile Edge Computing}
\acrodef{FL}{Federated Learning}

\begin{document}
\let\WriteBookmarks\relax
\def\floatpagepagefraction{1}
\def\textpagefraction{.001}

\shorttitle{Adaptive Client Selection with Personalization for Communication Efficient Federated Learning}

\shortauthors{Allan M. de Souza et~al.}

\title [mode = title]{Adaptive Client Selection with Personalization for Communication Efficient Federated Learning}                      



\author[1]{Allan M.\ {de Souza}}[orcid=0000-0001-7511-2910]
\cormark[1]
\ead{allanms@ic.unicamp.br}

\author[1,4]{Filipe Maciel}[orcid=0000-0002-9965-5124]
\ead{filipe.mr@ufc.br}

\author[1]{Joahannes B.\ D.\ {da Costa}}[orcid=0000-0001-9973-2479]
\ead{joahannes.costa@ic.unicamp.br}

\author[1]{Luiz F.\ Bittencourt}[orcid=0000-0001-6305-9059]
\ead{bit@unicamp.br}

\author[2]{Eduardo Cerqueira}[orcid=0000-0003-2162-6523]
\ead{cerqueira@ufpa.br}

\author[3]{Antonio A.\ F.\ Loureiro}[orcid=0000-0002-5250-1785]
\ead{loureiro@dcc.ufmg.br}

\author[1]{Leandro A.\ Villas}[orcid=0000-0002-3372-3366]
\ead{lvillas@unicamp.br}


\affiliation[1]{organization={Universidade Estadual de Campinas}, 
    city={Campinas},
    country={Brazil}
    }

\affiliation[2]{organization={Federal University of Pará}, 
    city={Belém},
    country={Brazil}}

\affiliation[3]{organization={Federal University of Minas Gerais}, 
    city={Belo Horizonte},
    country={Brazil}}

\affiliation[4]{organization={Federal University of Ceará},
            city={Russas}, 
            country={Brazil}}

\cortext[cor1]{Corresponding author}



\begin{abstract}
Federated Learning (FL) is a distributed approach to collaboratively training machine learning models. FL requires a high level of communication between the devices and a central server, thus imposing several challenges, including communication bottlenecks and network scalability. This article introduces \NAME\footnote{\url{https://github.com/AllanMSouza/ACSP-FL}}, a solution to reduce the overall communication and computation costs for training a model in  FL environments. \NAME~employs a client selection strategy that dynamically adapts the number of devices training the model and the number of rounds required to achieve convergence. Moreover, \NAME\  enables model personalization to improve clients performance. A use case based on human activity recognition datasets aims to show the impact and benefits of \NAME~ when compared to state-of-the-art approaches. Experimental evaluations show that \NAME~ minimizes the overall communication and computation overheads to train a model and converges the system efficiently. In particular, \NAME~reduces communication up to 95\% compared to literature approaches while providing good convergence even in scenarios where data is distributed differently, non-independent and identical way between client devices.
\end{abstract}




\makeatletter\def\Hy@Warning#1{}\makeatother

\maketitle

\section{Introduction}
Advances in distributed computing, such as the popularization of Edge Computing, Fog Computing, and, in a more general sense, \ac{MEC}~\cite{porambage2018survey, mec_survey}, have paved the way for distributed Machine Learning (ML) model training, utilizing the resources of end devices (i.e., devices located at the network edge)~\cite{mec_fl}. One of the techniques that leverages edge computing for distributed model training is called \ac{FL}~\cite{mec_fl, fl_1}.

\ac{FL} allows distributed model training, where it enables devices (i.e., clients) to fit a global model in their local data and then shares the trained model with a central server responsible for aggregating received models  (i.e., aggregator). Finally, the model is updated and shared again with clients for retraining. This whole procedure is known as communication round, repeating until the global model reaches a convergence. In other words, FL allows distributed devices to train collaboratively a shared model by keeping all training data on the device and sharing only the locally trained model. Consequently, it is possible to mitigate many systemic privacy risks and costs resulting from traditional, centralized machine learning approaches~\cite{ZHANG2020242}.

The model sharing among clients and the server in \ac{FL} solutions can introduce additional overhead to the system~\cite{survey_fl}. Therefore, they must consider the communication cost as an essential factor in the system design. Some variables influence the federated learning communication overhead, including: \textit{(i)}~the size of the shared model; \textit{(ii)}~the number of communication rounds to achieve model convergence; \textit{(iii)}~the number of clients selected per communication round; and \textit{(iv)}~the proportion of the shared model ranging from full or partial model sharing (i.e., few layers or subset of parameters). In this scenario, a still open research question is the following: how to reduce communication overhead without degrading the efficiency of the trained model?

The main approaches for reducing communication overhead focus on client selection and compression of the shared model~\cite{survey_fl}. However, these approaches must be more adaptive to explore the heterogeneity of devices and changes in the context of the environment. For example, \textit{Power-Of-Choice Selection}~\cite{cho_2020} defines a fixed number of clients for all communication rounds. At the same time, compression solutions use the same settings for all clients during all communication rounds. Other personalization schemes use a local model and a shared one in each client to customize and fine-tune the client model~\cite{personalized_fl}. Therefore, such approach enables partial model sharing, reducing the number of parameters transferred between clients and the server. However, as only some pieces of the model are shared, the client selection should be efficient to allow the system convergence accordingly.

This article presents {\NAME} (\COMPLETENAME) to allow an efficient networking communication procedure in FL environments. {\NAME} selects clients adaptively based on the global model performance to accelerate the model convergence while reducing communication and computation overheads during the training phase. Furthermore, {\NAME} reduces the size of the shared model in the uplink (i.e., from clients to the server) and the downlink (i.e., from server to clients) based on a partial model sharing. Finally, to not degrade model performance, \NAME\ also implements a personalization method, empowering clients to keep high performance even sharing only some layers of the model. The performance results  show that {\NAME} reduces the communication overhead by up to $90\%$ compared to existing solutions in the literature.

This article extends our previous work \cite{deev} by introducing new methods to improve both communication and computation aspects in FL environments, where it includes: \textit{(i)}~a partial model sharing between a set of clients and a server; and \textit{(ii)}~a personalization method to build and fine-tune the model in each client considering local and global models. In addition, this article also presents a detailed description of {\NAME}, including algorithms, performance evaluation environments, and the behavior and impact of the proposed solution when analyzed with different datasets. The main contributions of this article can be summarized as follows:

\begin{itemize}
    \item An efficient adaptive client selection approach for FL environments to reduce the communication overhead.

    \item A personalization mechanism combined with partial model training to reduce the communication overhead without degrading the model's performance.
    
    \item The description and implementation of an FL environment\footnote{\url{https://github.com/AllanMSouza/ACSP-FL}} (based on containers) for conducting experiments, separating client and server applications into different virtual machines. This division allows the creation of more heterogeneous, manageable, and reproducible environments for experimentation.
    
    \item The performance evaluation and assessed metrics to evaluate the proposed solution with literature approaches considering three different datasets. The results outline the benefits of an adaptive client selection scheme with personalization support to enhance the communication process in FL environments compared to other state-of-the-art approaches.
\end{itemize}

The remainder of this article is organized as follows. Section~\ref{sec:related} presents related works, highlighting their advantages and limitations. Section~\ref{sec:proposal} describes the proposed solution, including the methods for adaptive client selection, personalization and partial model sharing. Section~\ref{sec:evaluation} discusses the performance evaluation and results obtained. Finally, Section~\ref{sec:conclusion} concludes the work and presents some final remarks.

\section{Related Work}
\label{sec:related}

This section describes existing solutions that deal with communication overhead in FL environments, considering the following techniques: \textit{(i)}~model compression and sparsification; \textit{(ii)}~client selection; and \textit{(iii)}~personalization and partial model sharing.

The Federated Averaging (FedAvg) \cite{McMahan_2017} proposal reduced the number of communication rounds between clients and the central server by exploiting client processing in model training. Clients use several epochs, up to $100$, to update the local model before sending it to the server, consequently increasing the system's latency. However, it is not suitable for scenarios non-IID scenarios. 

One way to reduce the communication overhead in FL solutions is reducing the representation of the shared model with compression techniques~\cite{Shah_2022}, including quantization \cite{Lin_2018, Amiri_2020, Bernstein_2018} and sparsification \cite{Shah_2022, sattler_2019, Strom_2015}, in which a subset of parameters \cite{Shahid_2021}. ~\citet{Strom_2015} presents an approach that only sends parameters based on a threshold. \citet{Shah_2022} proposes a framework for developing compact models, where each client trains a sparse model locally and then communicates it to the server. Nevertheless, the aggregated model might no longer be sparse after federated averaging. Then, the central server performs a sparse reconstruction of the global model to recover some sparsity loss. There are two advantages to this operation: \textit{(i)}~it decreases the volume of downstream communication from the server to the clients, and \textit{(ii)}~it allows clients to train a sparse model instead of a dense one, reducing the processing overhead.

On the other hand, Amiri \textit{et al.} showed that quantization techniques increase communication efficiency \cite{Amiri_2020}. Thus, the central server uses the \textit{Lossy FL} (LFL) algorithm to transmit the quantized global model to clients, reducing the communication overhead without a performance decrease. Bernstein \textit{et al.} created SignSGD that only transmits the signal of each mini-batch stochastic gradient to alleviate the gradient communication bottleneck problem \cite{Bernstein_2018}. However, SignSGD does not perform well in scenarios with non-IID data.

Another way to reduce the communication overhead is the client selection. It selects a subset of available clients to train the shared model in a communication round. Consequently, this reduced participation diminishes the overall communication overhead. Such an approach also proved helpful in accelerating model convergence, reducing the number of rounds required for training the shared model~\cite{survey_fl, CS-Snapshot}.

\citet{McMahan_2017} introduced a random sampling client selection with FedAvg. However, the literature showed that by applying other methods to sampling clients, it is possible to improve a model performance and communication overhead~\cite{survey_fl}. \citet{NishioYonetani2019} observed the importance of systemic aspects, notably the training delay, for prioritizing clients for selection. \citet{goetz2019active} already considered statistical metrics about client training to guide the selection. \citet{Kang2019} merged both views to classify clients in terms of reputation and thus direct the selection process for the next round of training. 

\citet{cho_2020} proposed Power-Of-Choice (POC), a computationally and communication-efficient client selection framework that directs the process toward clients with the highest local loss. This selection allows faster convergence, increasing communication efficiency in heterogeneous FL environments. \textcolor{black}{Similarly, the proposal called Oort, created by \citet{oort}, utilizes loss to determine every client's training utility. However, they added training delay as a weight to consider and designed a more scalable selection algorithm than POC. Both algorithms define} a fixed number of clients (i.e., $k$) to train the global model in all rounds. This lack of adaptation according to the model's performance can provide redundant selections and even unnecessary even after model convergence.

\citet{personalized_fl} showed that there is poor convergence in scenarios with high data heterogeneity due to the problem of client drift, in which local models converge to a local optimum, degrading the average global model, which does not generalize to clients whose data distribution is quite different from the global one. For these scenarios, a proposed solution is to personalize the models for each client and thus achieve the desired performance. The literature describes two strategies: personalization of the global model and learning of personalized models. Personalizing the global model requires additional training for each client so that the global model adapts to the local data distribution. Personalized model learning modifies the aggregation process to allow each client to train a model that suits them.

Several scenarios characterized as non-independent and identically distributed use personalization with success \cite{personalized_fl}. For example, \citet{Chen2022} developed a method where each client can automatically balance the training of its local and global models, implicitly contributing to the training of other clients. Its Self-FL solution was applied to several datasets and non-iid scenarios, demonstrating superior performance to the baseline solutions.

Another fundamental aspect of saving resources is the optimization of the hyperparameters that determine the training behavior. \citet{Kairouz_2019} identified that one of the challenges for the learning system is identifying the minimum number of participants needed for each round of training for the model to converge. \citet{CE-FL} proposed an adaptive federated learning design, where the values of the training control variables are chosen to minimize the total cost and ensure convergence. One of these variables is the minimum number of participants per training round. Although the proposed algorithm determines a configuration that minimizes costs with little work overhead, its hyper-parameterization increases the complexity of the system that uses it.

Considering an adaptive client selection, ~\citet{deev} introduced DEEV, which samples clients based on their performance and applies a function to dynamically reduce the number of clients selected as the model reaches the convergence. This way, DEEV reduces the communication overhead produced by the federated learning process and also enables a faster model convergence. However, it does not implement any personalization mechanism to ensure a good performance for clients in non-IID scenarios.

\begin{table*}[ht]
\centering
\caption{Description of related works considering client selection, communication, and personalization methods.}\label{tab:relatedworks}
\scalebox{0.87}{
\begin{NiceTabular}{l|cccccc|ccc|ccc}
\toprule
\multirow{6}{*}{Related Work} & \multicolumn{6}{c}{\textbf{Client Selection}}                    & \multicolumn{3}{c}{\textbf{Communication}} & \multicolumn{3}{c}{\textbf{Personalization}} \\
\cmidrule{2-13}
                              & \rotatebox{65}{Random} & \rotatebox{65}{Statistical} & \rotatebox{65}{Systemic} & \rotatebox{65}{Performance} & \rotatebox{65}{Static} &\rotatebox{65}{Dynamic} & \rotatebox{65}{Quantization}  & \rotatebox{65}{Sparsification} & \rotatebox{65}{Partial Model} & \rotatebox{65}{Fine Tuning} & \rotatebox{65}{Layer Sharing} & \rotatebox{65}{Local Model}    \\
\midrule
\citet{McMahan_2017}                 & \checkmark &             &             &             &\checkmark &            &            &            &               &             &            &            \\
\citet{Lin_2018}                     &            &             &             &             &           &            & \checkmark &            &               &             &            &            \\
\citet{Amiri_2020}                   &            &             &             &             &           &            & \checkmark &            &               &             &            &            \\
\citet{Bernstein_2018}               &            &             &             &             &           &            & \checkmark &            &               &             &            &            \\
\citet{Shah_2022}                    &            &             &             &             &           &            &            & \checkmark &               &             &            &            \\
\citet{sattler_2019}                 &            &             &             &             &           &            &            & \checkmark &               &             &            &            \\
\citet{Strom_2015}                   &            &             &             &             &           &            &            & \checkmark &               &             &            &            \\
\citet{NishioYonetani2019}           &            &             & \checkmark  &             &\checkmark &            &            &            &               &             &            &            \\
\citet{goetz2019active}              &            & \checkmark  &             &             &\checkmark &            &            &            &               &             &            &            \\
\citet{Kang2019}                     &            & \checkmark  & \checkmark  &             &\checkmark &            &            &            &               &             &            &            \\
\citet{cho_2020}                     &            & \checkmark            &             & \checkmark  &\checkmark &            &            &            &               &             &            &            \\
\citet{ccs}                          &            & \checkmark            &             & \checkmark  &\checkmark &            &            & \checkmark &               &             &            &            \\
\citet{personalized_fl}              &            &             &             &             &\checkmark &            &            &            &               & \checkmark  &            &            \\
\citet{deev}                         &            & \checkmark             &             & \checkmark  &           &\checkmark  &            &            &               &             &            &            \\
\citet{rawcs}                        &            & \checkmark  & \checkmark  & \checkmark  &\checkmark &            &            &            &               &             &            &            \\
\citet{fedpredict}                   &            &             &             &             &\checkmark &            &            &            &               & \checkmark  &            & \checkmark \\
\textcolor{black}{\citet{oort}}                   &            & \checkmark             & \checkmark             & \checkmark             &\checkmark &            &            &            &               &   &            &  \\
\textbf{\NAME}                       &            &  \checkmark           &             & \checkmark  &           &\checkmark  &            &            & \checkmark    & \checkmark  &
\\\bottomrule
\end{NiceTabular}
}
\end{table*}

Motivated by the solutions presented in the literature and their limitations, we propose \NAME~, an approach for adaptive client selection that reduces the communication overhead in federated learning environments while enabling model personalization. Furthermore, \NAME~does not define a fixed parameter of clients to be selected and adjusts the number of participants according to the model's performance. \Cref{tab:relatedworks} summarizes the main characteristics of reviewed studies considering system design.   

\section{\NAME}
\label{sec:proposal}

\NAME\footnote{\url{https://github.com/AllanMSouza/ACSP-FL}}~aims to reduce the communication and processing overheads generated by FL solutions without degrading the efficiency of machine learning models. To achieve this, \NAME~ performs an adaptive selection of clients according to the model's performance to decide which devices should perform training in the next round of the communication process, consequently allowing the system to reduce its \textit{overhead} and increase its  performance. Moreover, \NAME~ also implements a personalization mechanism to improve the performance of the models in each client. Lastly, to reduce further the communication cost, \NAME~shares only selected layers of the global model to enable collaborative training cost-effectively.

\subsection{Problem Definition}

Let $|C|$ be the number of clients in a scenario where ${C}$ represents the set of clients, and $T$ is the number of communication rounds. In each round $t \in \{1, 2, 3, \ldots, T\}$, a selected subset $\mathcal{S} \subseteq {C}$ of clients takes part in training the model. Each client $i \in \mathcal{S}$ has its local dataset $d_i$, and $\mathcal{D}=\bigcup_{i \in C} d_i$. Thus, each client receives the current model parameters (i.e., weights) $w(t)$ from the server and updates its local model to train on its data. Subsequently, each client $i$ sends the newly trained model parameters back to the server after training $w^i(t+1)$. The server then aggregates the received models by performing a weighted average, which produces the updated model parameters for the next communication round. At the end of round $t$, the updated global $w(t+1)$ is computed by averaging the trained model $w^i(t+1)$  of each client $i$, according to \Cref{eq:model_agg}.
\begin{equation}
\label{eq:model_agg}
w^{(t+1)} = \sum_{i \in C}\frac{d_i}{|\mathcal{D}|}w_i^{(t)}
\end{equation}

Hence, the objective is to minimize \Cref{eq:modelUpdate}, where $w$ is the global model, $\mathcal{L}$ represents the model's loss function, and $x_i$ and $y_i$ are the input and label datasets, respectively.
\begin{equation}
\min f(w) \text{ where } f(w)=\sum_{i=1}^{|C|}\frac{d_i}{|\mathcal{D}|}\mathcal{L}(w; x_i, y_i)
\label{eq:modelUpdate}
\end{equation}

FedAvg \cite{McMahan_2017} solves \Cref{eq:modelUpdate} by dividing the training into communication rounds and randomly selecting a set of clients $\mathcal{S} \subseteq C$ to train the model. Each client trains the model on its local dataset with $\tau$ iterations (i.e., epochs) using the Stochastic Gradient Descent algorithm (SGD)~\cite{haykin1994neural} and sends the trained model back to the server. This way, the server updates the global model and shares it again with a new set of clients.

Formally, the set of $0 < i \leq |C|$ clients selected by the server is represented by $\mathcal{S}^{(t)}$. This is because the clients execute $\tau$ local iterations and the set of clients remains constant for each iteration. Therefore, the model update occurs as follows:
\begin{equation}
\footnotesize
    w^{(t+1)} = \displaystyle\frac{1}{|\mathcal{S}^{}|} \displaystyle\sum_{i \in \mathcal{S}^{}}(w^{(t)}_i - \eta_t \mathcal{L}(w_i^{(t)}, n_i)) 
\end{equation}
where $w_i^{(t + 1)}$ represents the local model of client $i$ at iteration $t$, $\eta$ is the learning rate and $\mathcal{L}(w_i^{(t) }, n_i)$ represents the SGD over the data fraction $n$ of client $i$. Additionally, $w^{(t+1)}$ defines the global model on the server. Although $w^{(t)}$ is only updated every $\tau$ iteration, for convergence reasons, we consider a sequence of $w^{(t)}$ that is updated every iteration. More details on the convergence of models with a partial client selection approach are presented in detail in~\cite{cho_2020}.

\subsection{Performance-based Client Selection}

Training models by performance-constrained clients allows models to achieve faster convergence and better average performance of federated learning-based solutions ~\cite{cho_2020}. Thus, clients who already have a good performance wait for clients with lower performance to train their models for more epochs and, consequently, their performance reaches the desired levels. In other words, this approach achieves a faster convergence towards the global average.

To do this, in addition to sharing the model, each client must share a metric with the server so that it is possible to measure the model's performance on its local data, such as training loss or even accuracy. Thus, the server is aware of each client's performance and can apply policies to allow the selection process based on the system performance. Such approaches were used to develop \NAME.
\begin{figure*}[!ht]
    \centering
    \subfigure[Personalization phase]{
        \label{fig:passo2}
        \includegraphics[width=0.31\textwidth]{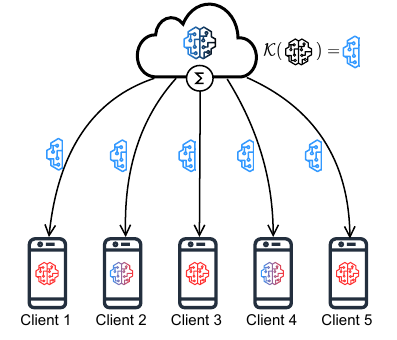}
    }
    \subfigure[Evaluation phase]{
        \label{fig:passo3}
        \includegraphics[width=0.31\textwidth]{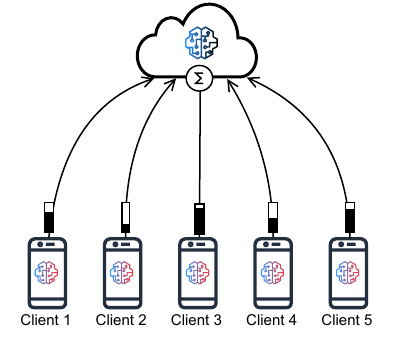}
    }
    \subfigure[Selection phase]{
        \label{fig:passo1}
        \includegraphics[width=0.31\textwidth]{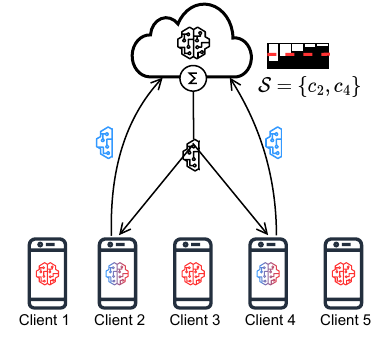}
    }
    
    \caption{Performance-based client selection with personalization: personalization phase (a) the server sends the model to the clients, the portion of the model is defined by $\mathcal{K}(\cdot)$; in evaluation phase (b) clients combine the layers shared by the server with their local model to train it with their local data and send back the performance metric to the server; selection phase (c), the server classifies clients according to performance and selects clients with lower performance than the average of received accuracies $\mathcal{A}$ to train the model in the next round of communication.}\label{fig:client-selection}
\end{figure*}

Figure~\ref{fig:client-selection} describes the client selection process applied by \NAME. The overall procedure employed by \NAME\ is organized as: \textit{(i)}~personalization phase; \textit{(ii)}~evaluation phase; and \textit{(iii)}~client selection phase. The personalization phase is described in \Cref{fig:passo2}, in which the model aggregated in the last round is sent to the clients for the following training procedure. However, \NAME~ does not send the entire model to the clients, and it sends just some layers defined by the function $\mathcal{K}(\cdot)$ (detailed described in the next section). Thus, when clients receive the shared piece, they combine it with its local model for personalization. Hereafter starts the evaluate phase, described by \Cref{fig:passo3}, where the clients evaluate the model personalized in its test data and report the performance metric to the server. Finally, when the server receives the performance metric of each client, it defines the set of clients that will be selected to train the model in the following communication round based on the adaptive client selection based on performance. In this way, the selected clients receive the shared piece of the model, personalize it with its local model, fit the model in its local data, and port the shared trained piece to the server. Thus, the described sequence repeats until convergence.

\textcolor{black}{In particular, \NAME\ selects clients considering their accuracy ($\mathcal{A}_i$). The function $\pi(\cdot)$ is applied as a filtering procedure, so each client that fits the filtering is selected to be part of the set $\mathcal{S}$. The filtering mechanism verifies whether the performance of client $i \in C$ is lower or equal to the mean performance given by the following equation:
\begin{equation}
\pi(i,\mathcal{A}) = \begin{cases}
    i & \text{ if } \mathcal{A}_i \leq \mathcal{A}_{mean}\\
    \emptyset & \textit{otherwise}
\end{cases}
\end{equation}
where $\mathcal{A}$ is the set of accuracies of each client. $\mathcal{A}_i$ and $\mathcal{A}_{mean}$ are the accuracy of the client $i \in C$ and the average accuracy, respectively. Then, by applying the filtering function to all clients in the system, \NAME~ defines the set of selected clients, represented by: 
\begin{equation}
\label{eq:filtering_clients}
\mathcal{S} = \displaystyle\bigcup_{i \in C} \pi(i, \mathcal{A}) 
\end{equation}
where $\pi(\cdot)$ is the filtering method that considers the performance of each client $i \in C$. This approach reduces the communication overhead introduced by federated learning, as fewer clients share models each round, enabling faster convergence. Furthermore, it is important to highlight that the selection procedure implemented by \NAME~does not require defining the number of clients a priori, as it is dynamically adjusted at each communication round based on the average performance of the customers.}

\subsection{Adaptive Client Selection}

Adjusting client selection according to its performance is essential to reduce the use of communication and processing resources. An obstacle to this is the fixed number of selected clients (i.e., $k$-clients) in each round, even after the model convergence, where it could decrease. Another problem is that selection algorithms might include clients that already perform well and do not contribute to improving the model's convergence just to complete the defined number of $k$-clients, incurring unnecessary communication and processing.

\begin{figure}[!htbp]
    \centering
    \includegraphics[width=0.45\textwidth]{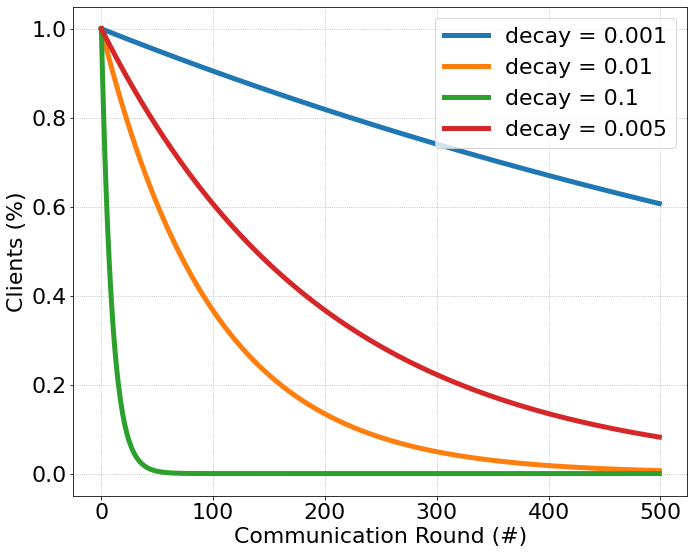}
    \caption{Example of the decay mechanism employed by \NAME.}
    \label{fig:dacay}
\end{figure}

\NAME~implements a gradual reduction of clients so that fewer customers are selected in each round based on model's performance. Consequently, further reducing the use of communication and processing resources. The gradual reduction in the number of clients is implemented by a decay function presented in Equation~\ref{eq:gradualReduction} as follows:
\begin{equation}
\centering
    \phi(\mathcal{S}, t) = \ceil{\,|\mathcal{S}| \cdot (1 - \textit{decay})^t\,}
\label{eq:gradualReduction}
\end{equation}
where $\mathcal{S}$ represents the set of clients previously selected by \NAME~in round $t$ and decay is a parameter to define how gradual the decline in the number of clients will be. Figure~\ref{fig:dacay} illustrates the gradual decay of clients according to the value of the decay parameter. It is important to highlight that this decay is carried out after defining the adaptive client set implemented by \NAME. As can be seen, different values provide different decays, which must be adjusted to allow a better cost-benefit ratio between performance and overhead.

With the number of clients calculated after defining the set of clients that perform less than the average performance (i.e., $\mathcal{A}_{mean}$, given by \Cref{eq:filtering_clients}) and also after applying the decay function, \NAME~selects the clients as follows:
\begin{equation}
    \mathcal{S} = \bigcup_{i=0}^{\phi(\mathcal{S}, t)} \mathcal{S}[i], \text{ such that } \phi(\mathcal{S}, t) \leq |\mathcal{S}|
\end{equation}
that is, after ordering the clients' performance \NAME~ selects the $\phi(\mathcal{S}, t)$ first clients with performance below the average.

\subsection{Model Personalization and Layer Sharing}
\color{black}
Personalization empowers model performance via fine tuning~\cite{personalized_fl}. The idea is to adjust the global model on clients data, thus providing a better fit for each client. The traditional approach for personalization and fine tuning is to keep one local model in each client (i.e., $w^l_i \ i \in C$, local model of client $i$) and a global model which is the aggregation of the local models of each client (i.e., $w^g = \sum_{i\in C}\frac{n_i}{N}w^l_i$). Therefore, with those models the client can decide which one will be used based on the one that achieves the best performance, according to the personalization function $\mathcal{P}(w^l_i, w^g_i)$:
\begin{equation}
\label{eq:personalization}
\mathcal{P}(w^l_i, w^g) = \begin{cases}
    w^l_i & \mathcal{L}(w^l_i; x_i, y_i) \leq \mathcal{L}(w^g; x_i, y_i)  \\
    w^g_i & \textit{otherwise}
\end{cases}
\end{equation}
where $w^l_i$ and $w^g_i$ are the local and the global model of the client $i$, $\mathcal{L}(w^l_i; x_i, y_i)$ is the performance of the local model and $\mathcal{L}(w^g; x_i, y_i)$ is the performance of the global model on clients data.

This approach might produce several overfitted models, which is not the desired behavior expected by the FL environments. Thus, to enhance the system performance while taking advantage of both approaches, \NAME~ divides the model of each client into two parts. One set of layers will be part of the collaboratively trained global model ($w^g$). The other layers will compose the personalized model ($w^l_i$) to be trained only locally on the client's data. Therefore, the model of the client $i \in C$ is a composition as $w_i = [w^g, w^l_i]$.

To define which layers will be trained cooperatively, \NAME~ uses a function $\mathcal{K}(w, L)$ to cut the model according to the layers $L = \{l_1, l_2, \ldots, l_m\} \in w$, where each layer $l_j \in L$ represents a layer of the global model. \NAME can define the number of layers statically, in which the same number is used for all clients in all communication rounds or dynamically based some algorithm or function considering the performance of a client in a particular round. 

\begin{figure}[!ht]
    \centering
    \includegraphics[width=0.4\textwidth]{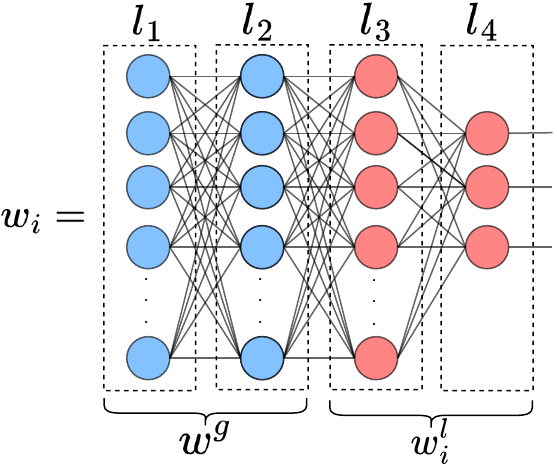}
    \caption{Model construction employed by \NAME~ based on the shared and local pieces.}
    \label{fig:model_personalization}
\end{figure}

\Cref{fig:model_personalization} shows how \NAME~ builds the model in each client combining the shared layers (i.e., the global piece $w^g$) with the local layers personalized in each client (i.e., the local piece $w^l_i$).  In this example,  $L = \{l_1, l_2\}$ defines that the first two layers, the black neurons, are the shared global model $w^g = \mathcal{K}(w, L)$, while the remaining part, the red neurons, will be personalized on its local data. Therefore, combining these pieces together each client builds its own model $w_i = [w^g, w^l_i]$.

Eventually, layer sharing among clients enables better performance and reduces communication overhead in both uplink and downlink with fewer parameters transmission over the network.

\normalcolor

\subsection{Proposed Algorithms }

This section describes the procedures employed by the server and the clients with \NAME. This way, Algorithm~\ref{alg:server} shows how the server selects the clients for each round and how it defines the pieces of the model that will be shared, while Algorithm~\ref{alg:client} presents how the clients personalize the model and fit it on its local data and also how they evaluate the trained model and report its performance for the server. 

In particular, the server initiates the procedure defining the number of clients participating in the federated learning procedure $C$ and the number of communication rounds $T$. Then, it initializes the model $w$ and selects all clients for the first communication rounds (Lines~\ref{alg:model_init}-\ref{alg:all_clients}). Thus, for each communication and client selected (i.e., $i \in \mathcal{S}$) the server defines the shared layers among all clients (Line~\ref{alg:define_l}), cuts the model based on the set of layers defined (Line~\ref{alg:cut_model}) and sends it to each client to perform the local fit (Line~\ref{alg:send_clients}).

Hereafter, the server receives the trained pieces $w^g_i$ of each client and computes the total size of the dataset $\mathcal{D}$ based on the size reported by each client $d_i$ to perform the aggregation (Lines~\ref{alg:compute_d}-\ref{alg:aggregate_models}). Moreover, with the aggregated model, the server sends it to be evaluated by each client and to receive their performance metric (Line~\ref{alg:distributed_eval}). Then, based on the performance of each client, the server applies the adaptive client selection procedure to define the set of clients $\mathcal{S}$ for the next communication round (Line~\ref{alg:client_selection}). The entire procedure applied by \NAME~for the client selection is described from Line~\ref{alg:temp_set} to Line~\ref{alg:clients_next_round}. \begin{algorithm}[t]
\footnotesize
\caption{Server Function}\label{alg:server}
\SetKwInOut{Input}{input}
\SetKwInOut{Output}{output}
    \Input{ $C$ \tcc{Set of clients it the system}\\
             \nonl $T$ \tcc{Number of communication rounds}\\
    }
\SetKwProg{Fn}{Server procedure}{:}{}
\Fn{($C, T$)}{
    \tcp{initialize the model}
    $w \gets $\textsc{RandomInit(})\;\label{alg:model_init}
    \tcp{select all clients for the first round}
    $S \gets C$\;\label{alg:all_clients}
    \For{$t \in \{1, 2, 3, \ldots T\}$}{
        \ForEach{$i \in \mathcal{S}$}{
            \tcp{defines the part of the model that will be shared}
            $L   \gets$ \textsc{SharedLayers}($w$)\;\label{alg:define_l}
            \tcp{cut the model based on the layers $L$}
            $w^g \gets \mathcal{K}(w, L)$\;\label{alg:cut_model}
            \tcp{Send the shared part to clients $i \in \mathcal{S}$}
            $w_i, |d_i| \gets$ \textsc{LocalTrain}($i, w^g)$\;\label{alg:send_clients}
        }
        $\mathcal{D} \gets \sum_{i\in\mathcal{S}}d_i$\label{alg:compute_d}
        \tcp{aggregates the received models}
        $w   \gets \displaystyle\sum_{i \in \mathcal{S}}\frac{|d_i|}{\mathcal{D}}w_i$\;\label{alg:aggregate_models}
        \tcp{send the aggregated model for evaluation}
        $\mathcal{A} \gets$ \textsc{Evaluate(}$w, C$)\;\label{alg:distributed_eval}
        \tcp{compute the set of selected clients for the next round}
        $\mathcal{S}   \gets$ \textsc{ClientSelection(}$C, \mathcal{A}, t$)\;\label{alg:client_selection}
    }
}
\SetKwProg{Fn}{ClientSelection}{:}{}
\Fn{($C, \mathcal{A}$, t)}{
    $S' \gets []$\;\label{alg:temp_set}
    \For{$i \in C$}{
        \tcp{filter clients based on condition $\pi(\cdot)$}
        $S' \bigcup \pi(i, \mathcal{A})$\;\label{alg:filter_clients}
    }
    \tcp{filter clients based on the decay factor}
    $\mathcal{S} \gets \displaystyle\bigcup_{j=0}^{\phi(\mathcal{S'}, t)} \mathcal{S'}[j]$\; \label{alg:apply_decay}
    \tcp{return selected clients}
    \Return{$\mathcal{S}\;$}\label{alg:clients_next_round}
}
\end{algorithm}

\begin{algorithm}[t]
\footnotesize
\caption{Client training}\label{alg:client}
\SetKwInOut{Data}{data}
\Data{  $i$ \tcc{client identification}\\
    \nonl $d_i$ \tcc{local dataset of client $i$} \\
    \nonl $w^l_i$ \tcc{local model of client $i$}
}

\SetKwProg{Fn}{LocalTrain}{:}{}
\Fn{($i, w^g)$}{
    \tcp{combines the local model with the local one}
    $w_i \gets [w^g, w^l_i]$\;\label{alg:personalize_model}
    \For{$\text{epoch } \in \{1, 2, 3, \ldots, \tau\}$}{
        \tcp{split the train data in batches}
        \ForEach{$(x_i, y_i) \in d_i$}{\label{alg:split_batches}
            \tcp{compute the loss for each batch}
            $\mathcal{L}(w_i; x_i, y_i)$\;\label{alg:fit_model}
        }
    }
    \tcp{return trained piece and the size of $d_i$ }
    \Return{$w^g_i, |d_i|\;$}\label{alg:return_fited_model}
}
\SetKwProg{Fn}{Evaluate}{:}{}
\Fn{($w^g, i)$}{
    \tcp{combines the local model with the local one}
    $w_i \gets [w^g, w^l_i]$\;\label{alg:personalize_model_2}
    \tcp{evaluate the model on test data}
    $\mathcal{A}_i \gets $ \textsc{Eval(}$w_i, N_i$)\;\label{alg:evaluate_model}
    \tcp{return the performance for the server}
    \Return{$\mathcal{A}_i$}\label{alg:return_performance}
}
\end{algorithm}

On the other hand, when each selected client $i \in \mathcal{S}$ receives the local train request, it produces the model $w$ to be trained by combining the shared piece (i.e., global model $w^g$) with the local piece (i.e., $w^l_i$) described in Line~\ref{alg:personalize_model}. Then, for each local epoch, the client splits the local train dataset into batches ($x_i, y_i$) to fit the model (Lines~\ref{alg:split_batches}-\ref{alg:fit_model}). With the model trained, the client reports the shared piece $w^g_i$ to the server with the size of its train dataset $d_i$. These are the steps to train the model considering the personalization approach that combines the local model with the global one. It is essential to highlight that in such communication procedure (i.e., it sends the pieces of the model for the client and reports it back to the server) \NAME~reduces the communication overhead because fewer parameters are shared, saving bandwidth in both directions uplink and downlink, while keeping the system accuracy high.

Lastly, when clients receive the evaluation request from the server to perform the distributed evaluation of the model, each client also needs to build the model based on the local and shared pieces (Line~\ref{alg:personalize_model_2}) to evaluate it on its test data (Line~\ref{alg:evaluate_model}). This procedure returns the performance metric $\mathcal{A}_i$ used to evaluate the model's performance in each client. Then, this metric backs to the server (Line~\ref{alg:return_performance}), which will use it to perform the adaptive client selection method.

\section{Performance Evaluation}

\label{sec:evaluation}

This section presents the evaluation of {\NAME} against the state-of-the-art works FedAvg~\cite{McMahan_2017}, POC~\cite{cho_2020}, and DEEV~\cite{deev}, analyzing the model's performance, client selection, as well as communication and processing overhead.
For this purpose, we used a Docker Swarm to create multiple virtual machines (VMs) organized as either VM managers or VM workers.
Additionally, each application, namely, server and client, is deployed on the VMs as containers.
Each type of VM is responsible for running a specific type of application. Thus, the VM managers run the server application, while the VM workers run one or more client applications (i.e., containers).

\subsection{Evaluation Environment}
\label{subsec:ambiente}

A container-based environment (i.e., dockerized) runs the evaluation, where the main idea is to separate the server and client applications onto different virtual machines. To achieve this, we used a Docker Swarm to create multiple virtual machines (VMs), acting as VM managers or VM workers. Furthermore, each application, namely, server and client, is deployed on the VMs as containers. Each type of VM is responsible for executing a specific type of application. Therefore, the VM managers run the server application, while the VM workers run one or more client applications (i.e., containers). 

Client containers are responsible for storing local data, training the model locally with their data, compressing the model to enhance communication efficiency, sharing the trained model with the server for collaborative learning, and updating the model after the server performs aggregation.
Communication between the clients and the server is facilitated through gRPC using the Flower framework~\cite{beutel2020flower}.
On the other hand, the server application aggregates the models received from the clients, evaluates the overall model performance, selects a set of clients for training in the next round of communication, and ultimately provides communication logs, resource usage, and performance metrics. These logs are obtained using the Docker engine API.

A \textit{docker-compose} file sets up the environment where different parameters can be passed through each container using environment variables, including neural network configuration, aggregation approach, compression method, and resource allocation (i.e., CPU and RAM limits). It is worth noting that the network attribute in the \textit{docker-compose} file can define the usage of different communication technologies. Additionally, the customization of CPU and RAM limitations for each container can determine heterogeneous clients.

\subsection{Dataset, Application, and Model}
\label{subsec:dataset}
To evaluate the performance of the FL algorithms, we consider a Human Activity Recognition (HAR) application using sensor data collected by smartphones. In this evaluation, we considered three HAR datasets:
\begin{itemize}
    \item \textbf{UCI-HAR Dataset}: is a widely recognized benchmark for human activity recognition, featuring accelerometer and gyroscope data from smartphones worn by approximately 30 users. These users perform activities including walking upstairs, downstairs, sitting, standing, and lying. The dataset provides time-series sensor data and corresponding activity labels, serving as a fundamental resource for assessing and developing human activity recognition algorithms in the machine learning and data mining communities~\cite{uci-har}.
    
    \item \textbf{MotionSense Dataset}: is a human activity recognition containing sensor data from accelerometers and gyroscopes of smartphones worn by 24 users engaged in activities such as walking, jogging, sitting, standing, and climbing stairs. This dataset provides time-series sensor data across three dimensions ($x, y, z$) and corresponding activity labels, making it a robust tool for activity recognition and sensor data analysis research~\cite{Malekzadeh:2019:MSD:3302505.3310068}.

    \item \textbf{ExtraSensory Dataset}: is a unique dataset designed for mobile and wearable sensing research, comprising sensor data collected from smartphones and wearable devices used by 60 users. It encompasses data from various sensors, including accelerometers, GPS, light sensors, and more, along with rich contextual information such as location, time, and weather conditions. Participants engage in diverse activities, making this dataset invaluable for context-aware applications, activity recognition research, and exploring relationships between sensor data and user context~\cite{ExtraSensory}.
\end{itemize}

\begin{figure*}[ht]
    \centering
    \includegraphics[width=\textwidth]{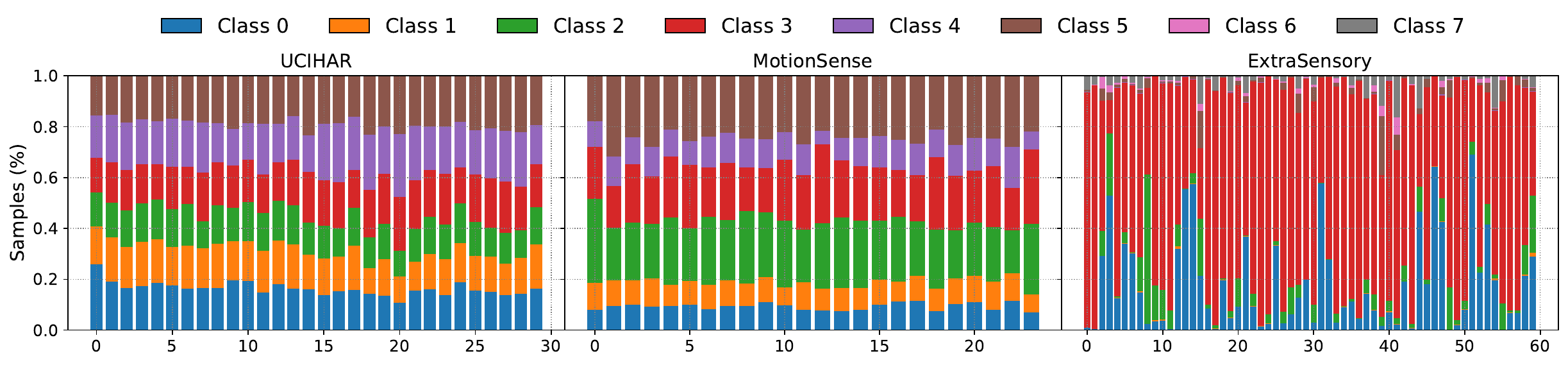}
    \caption{\textcolor{black}{Visualization of classes distribution by client for each dataset.}}
    \label{fig:datasets}
\end{figure*}

\textcolor{black}{\Cref{fig:datasets} displays the class distribution for each client across various datasets. As observed, the UCI-HAR and MotionSense datasets exhibit a comparable number of clients, as well as a similar distribution, suggesting identically distributed (IID) characteristics. In contrast, the ExtraSensory dataset features a larger number of clients, and the class distribution among these clients is non-identically distributed (Non-IID). \Cref{tab:datasets} describes more details for each dataset, including number of clients, average samples per client, number of features and also minimum and maximum number of samples in clients.}
\begin{table}[!hb]
\centering
\caption{Dataset characteristics.}\label{tab:datasets}
\resizebox{0.45\textwidth}{!}{%
\color{black}
\begin{tabular}{lccc}
\toprule
                     & \multicolumn{3}{c}{\textbf{Dataset}}                            \\ \midrule
                     & \textbf{UCI-HAR} & \textbf{MotionSense} & \textbf{ExtraSensory} \\ \midrule
$\#$ Clients         & 30               & 24                   & 60                    \\
$\#$ Classes         & 6                & 6                    & 8                     \\
$\#$ Features        & 561              & 7                    & 277                   \\
Samples per client (avg.)  & 274              & 47095                & 5015                  \\
Client with min. $\#$ samples          & 224              & 40804                & 1280                  \\
Client with max. $\#$ samples          & 327              & 57559                & 9596                 \\
\bottomrule
\end{tabular}%
\normalcolor
}
\end{table}

A suitable model for the HAR problem is the \textit{MultiLayer Perceptron} (MLP) \cite{haykin1994neural} composed of three hidden layers, with 256 units per hidden layer~\cite{8090454}. The following parameters adjust the MLP: \textit{Stochastic Gradient Descent} (SGD) as a method to update the weights of the MLP model and \textit{Sparse Categorical Crossentropy} as a loss function. We want to optimize the accuracy score as it is a classification task. The objective is to train the global model (MLP) on local customer data, evaluate it in a distributed way, and achieve good performance.

\subsection{Assessed Metrics}\label{subsec:metricas}

To evaluate the performance and also the communication and processing overhead of the solutions, the following metrics were defined:

\begin{itemize}
    \item \textbf{Distributed accuracy}: measures the convergence of the model. Thus, the accuracy is distributively computed in each communication round. Each client evaluates the aggregated model in its local test dataset and reports the result to the central server, which computes the average accuracy considering all clients' reports. The fewer rounds needed to reach the higher accuracy, the faster the solution will converge.

    \item \textbf{Overhead}: measures the time spent in each round for each solution, considering: \textit{(i)} the time to receive the model from the server, \textit{(ii)} the processing time to train the model, and \textit{(iii)} time to transmit the trained model back to the server. To compute this metric, we define the FedAvg's latency as the baseline. Thus, the results represent the percentage of reduction compared to the latency of FedAvg.

    \item \textcolor{black}{\textbf{Data transmitted}: this metric evaluates the data transmitted, specifically the TX bytes, taking into account both the overall data transmitted and the data transmitted per client. To calculate this metric, we utilized the size of the model sent by each client in bytes during each round.}

    \item \textcolor{black}{\textbf{Convergence time}: shows the total time, in seconds, taken by each solution to complete 100 communication rounds. In essence, it quantifies the time required for each solution to conclude its execution.}

    \item \textcolor{black}{\textbf{Efficiency}: this metric evaluates the efficiency of the solution to reach the provided accuracy by considering a weighted sum of the average accuracy and the overhead reduction. In particular, the weighted sum is calculated as $\alpha \times \mathcal{A}_{mean} + \beta \times \textit{overhead}\ \textit{reduction}$. These parameters can be adjusted according to the importance of the accuracy and the communication, but in this evaluation we set the value of both parameters to $0.5$.}

    \item \textbf{Accuracy distribution}: shows the final test accuracy in each client in the last communication round. Thus, we can show whether all clients had good performance or some clients with high performance and others with poor ones. In other words, with this metric, we want to see how homogeneous the performance of all clients is.
    
    \item \textbf{Client selection}: this metric assesses how often the clients were selected for the adaptive client selection solutions. The higher number of clients means higher communication overhead. The key idea is to understand the average number of times clients participated in the training.
    
\end{itemize}

\subsection{\NAME\ Analysis}

\begin{figure*}[!h]
    \centering
    \includegraphics[width=0.99\textwidth]{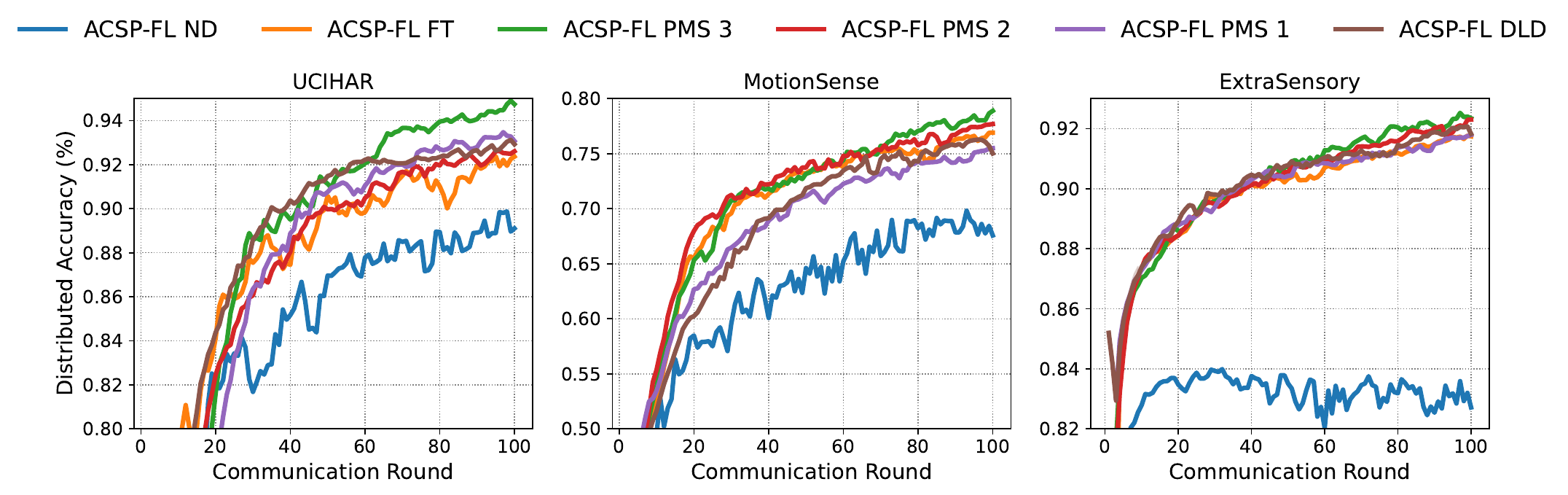}
    \caption{Test accuracy in function of the communication round for each dataset.}
    \label{fig:acc_acpfl}
\end{figure*}

\textcolor{black}{In this section, we present an exploratory analysis of the methods implemented by \NAME\ to assess how personalization, partial model sharing, and the client selection mechanism impact its performance in terms of \textit{(i)}~accuracy; \textit{(ii)}~communication overhead; \textit{(iii)}~convergence time; and \textit{(iv)}~efficiency. \Cref{fig:acc_acpfl} illustrates the distributed accuracy for each evaluated dataset. For clarity, we have named the solutions based on the number of layers shared between clients and the server. It is essential to note that \NAME\ FT represents the fine-tuned approach, where each client decides whether to retain its local model or adopt the global one, with no partial model sharing, while \NAME\ PMS (\#) represents the number of layer of partial model sharing, \NAME\ DLD represents the dynamic layer definition function and \NAME\ ND represents the approach without using personalization and the decaying function.}

\begin{figure}[h]
    \centering
    \includegraphics[width=0.4\textwidth]{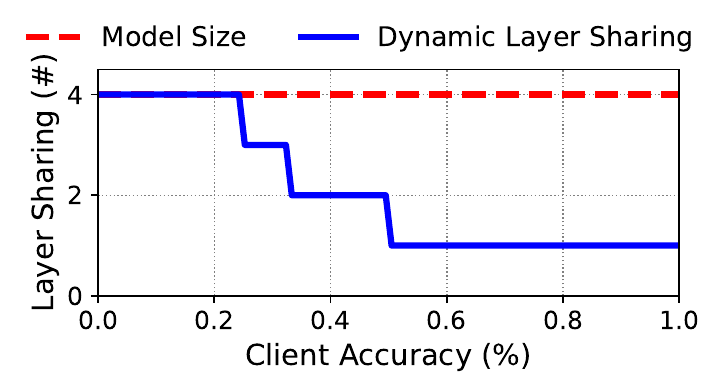}
    \caption{Example of dynamic layer definition function based on client's accuracy.}
    \label{fig:dld}
\end{figure}

\textcolor{black}{The dynamic layer definition employed by \NAME\ ND is given by:
\begin{equation}
    PMS = \begin{cases}
        4 & \mathcal{A}^t \leq 0.25 \\
        \big\lceil{\frac{1}{\mathcal{A}^t}}\big\rceil & \textit{otherwise}
    \end{cases}
\end{equation}
where 4 is the total number of layers of the model used in the evaluation and $\mathcal{A}^t$ is the accuracy of the client at the communication round $t$. Figure~\ref{fig:dld} shows an example of how to determine the number of layers to share based on the clients' accuracy, as the accuracy increases fewer layers are shared giving higher importance to the local model but also considering the knowledge of the last layer trained cooperatively.}

\begin{figure*}[!h]
    \centering
    \includegraphics[width=\textwidth]{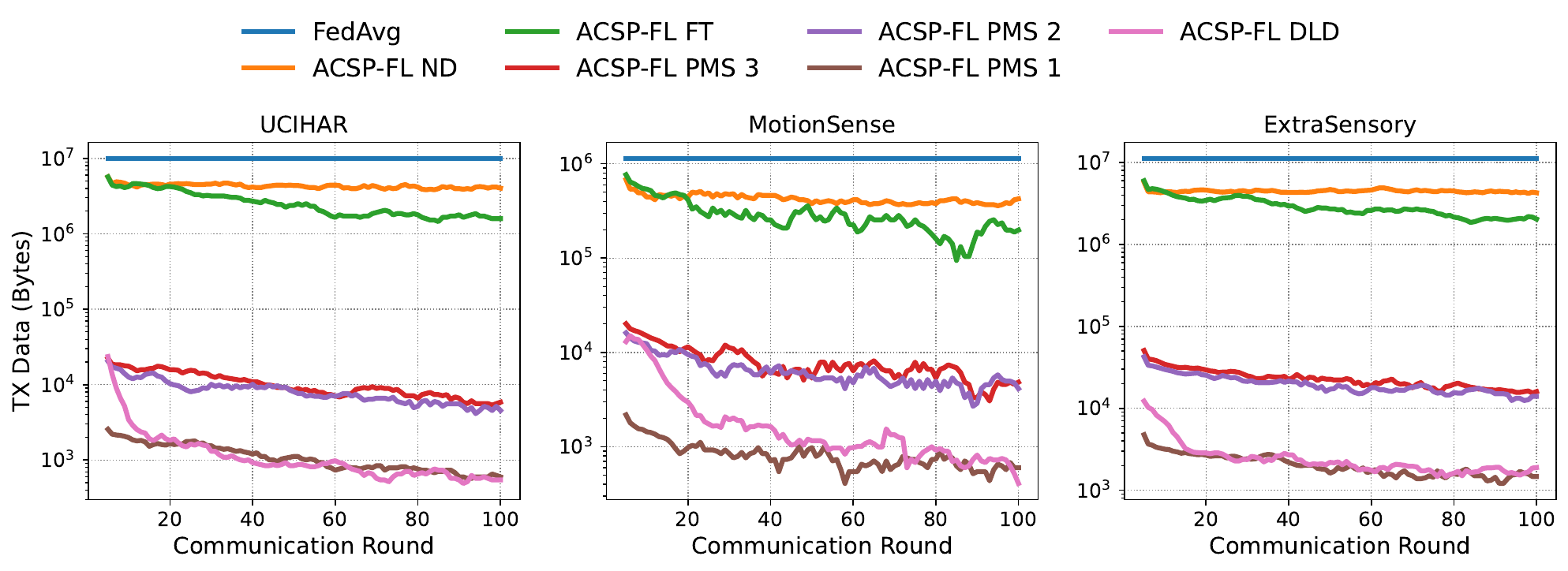}
    \caption{Analysis of communication overhead considering the partial model sharing and also the decaying function}
    \label{fig:pms_decay}
\end{figure*}

\textcolor{black}{The results indicate that, despite the number of layers shared, \NAME\ can still converge for each dataset (refer to \Cref{fig:acc_acpfl}). However, the \NAME\ ND variant cannot achieve the same accuracy as the other variations due to the absence of model personalization. On the other hand, \Cref{fig:pms_decay} illustrates the data transmitted by each solution, revealing three main aspects. Firstly, the \NAME\ variant that employs the decaying function reduces the data transferred as the training phase progresses, as fewer clients are selected when the model converges. In contrast, \NAME\ ND, which does not use the decaying function, experiences constant communication overhead across all datasets. The second aspect is that, with fewer layers shared by \NAME\ PMS variants, less data is transferred. Lastly, \NAME\ DLD, which determines the number of layers shared in each round by each client according to its accuracy, adapts itself to reduce communication overhead. Therefore, it proves to be a suitable solution, as no hyperparameter exploration is needed for different scenarios.}

\textcolor{black}{The performance of \NAME\ variants for other metrics is summarized in \Cref{tab:acsp-fl_variants}, which presents the performance of each variant considering accuracy, total data transmitted, average data transmitted per client, convergence time, and the efficiency metric. The results indicate that personalization enhances model accuracy, while partial model sharing reduces overhead by up to $99\%$. Additionally, as less data is transmitted, it also reduces the convergence time for variants that utilize partial model sharing. It is important to note that \NAME\ DLD slightly increases the convergence time due to the dynamic layer definition function that needs to be computed every single round. Finally, the efficiency metric shows that the combination of personalization and partial model sharing improves the efficiency of the solution. The closer the metric is to 1, the better the efficiency of the solution, indicating its ability to deliver a reduction in communication overhead similar to its accuracy.}


In summary, by combining these three approaches \textit{(i)}~adaptive client selection with the decaying function; \textit{(ii)}~model personalization; and \textit{(iii)}~partial model sharing, \NAME\ reaches higher accuracy level in early stages of the federated cycle. Empowered by the personalization method, fewer clients start to be selected due to the adaptive client selection, and in this whole cycle, less data is transmitted over the network due to the partial model sharing, demonstrating how efficient \NAME\ is.


\begin{table*}[]
\caption{Summary of assessed metrics for \NAME\ variations.}\label{tab:acsp-fl_variants}
\resizebox{\textwidth}{!}{%
\color{black}
\begin{tabular}{ccccccc}
\toprule
                                       &                                 & \multicolumn{5}{c}{\textbf{Metrics}}                                                                                    \\ \midrule
\textbf{}                              & \textbf{Solutions}              & \textbf{Accuracy (\%)}                    & \textbf{TX bytes (Mb)} & \textbf{TX bytes per client (Mb)} & \textbf{Convergence time (s)} & \textbf{Efficiency} \\ \midrule
\multirow{6}{*}{\textbf{UCI-HAR}}      & ACSP-FL ND                      & 0.90                      & 434               & 14.4                         & 19.14                & 0.71       \\
                                       & ACSP-FL FT                      & 0.89                      & 268               & 8.9                          & 18.15                & 0.79       \\
                                       & ACSP-FL PMS 3                   & 0.92                      & 1.07              & 0.03                         & 8.34                 & 0.96       \\
                                       & ACSP-FL PMS 2                   & 0.90                      & 0.85              & 0.02                         & 8.31                 & 0.95       \\
                                       & ACSP-FL PMS 1                   & 0.91                      & 0.11              & 0.003                        & 8.30                 & 0.95       \\
                                       & \multicolumn{1}{c}{ACSP-FL DLD} & \multicolumn{1}{c}{0.92 } & 0.22              & 0.007                        & 9.60                 & 0.95       \\ \midrule
\multirow{6}{*}{\textbf{MotionSense}}  & ACSP-FL ND                      & 0.70                      & 43.1              & 1.79                         & 221.27               & 0.67       \\
                                       & ACSP-FL FT                      & 0.71                       & 30.0              & 1.25                         & 221.27               & 0.74       \\
                                       & ACSP-FL PMS 3                   & 0.71                      & 0.82              & 0.03                         & 209.63               & 0.88       \\
                                       & ACSP-FL PMS 2                   & 0.71                      & 0.66              & 0.02                         & 208.13               & 0.87       \\
                                       & ACSP-FL PMS 1                   & 0.68                      & 0.08              & 0.003                        & 210.85               & 0.87       \\
                                       & \multicolumn{1}{l}{ACSP-FL DLD} & 0.70                      & 0.24              & 0.010                        & 216.76               & 0.87       \\ \midrule
\multirow{6}{*}{\textbf{ExtraSensory}} & ACSP-FL ND                      & 0.92                      & 452               & 7.53                         & 40.72                & 0.74       \\
                                       & ACSP-FL FT                      & 0.92                      & 298               & 4.97                         & 38.39                & 0.82       \\
                                       & ACSP-FL PMS 3                   & 0.95                      & 2.38              & 0.039                        & 31.48                & 0.95       \\
                                       & ACSP-FL PMS 2                   & 0.92                      & 2.03              & 0.033                        & 30.99                & 0.95       \\
                                       & ACSP-FL PMS 1                   & 0.92                      & 0.21              & 0.003                        & 30.77                & 0.95       \\
                                       & \multicolumn{1}{l}{ACSP-FL DLD} & 0.93                      & 0.28              & 0.004                        & 38.91                & 0.95      \\ \bottomrule
\end{tabular}%
}
\end{table*}
\normalcolor

\subsection{\NAME\ vs. Literature Analysis}
\label{subsec:resultados_acuracia}

\textcolor{black}{This evaluation aims to assess the federated learning efficiency of each solution: FedAvg~\cite{McMahan_2017}, POC~\cite{cho_2020}, DEEV~\cite{deev}, OORT~\cite{oort}, and {\NAME} with personalization and partial model sharing using the dynamic layer definition (i.e., \NAME\ DLD).} 

\begin{figure*}[ht]
    \centering
    \includegraphics[width=\textwidth]{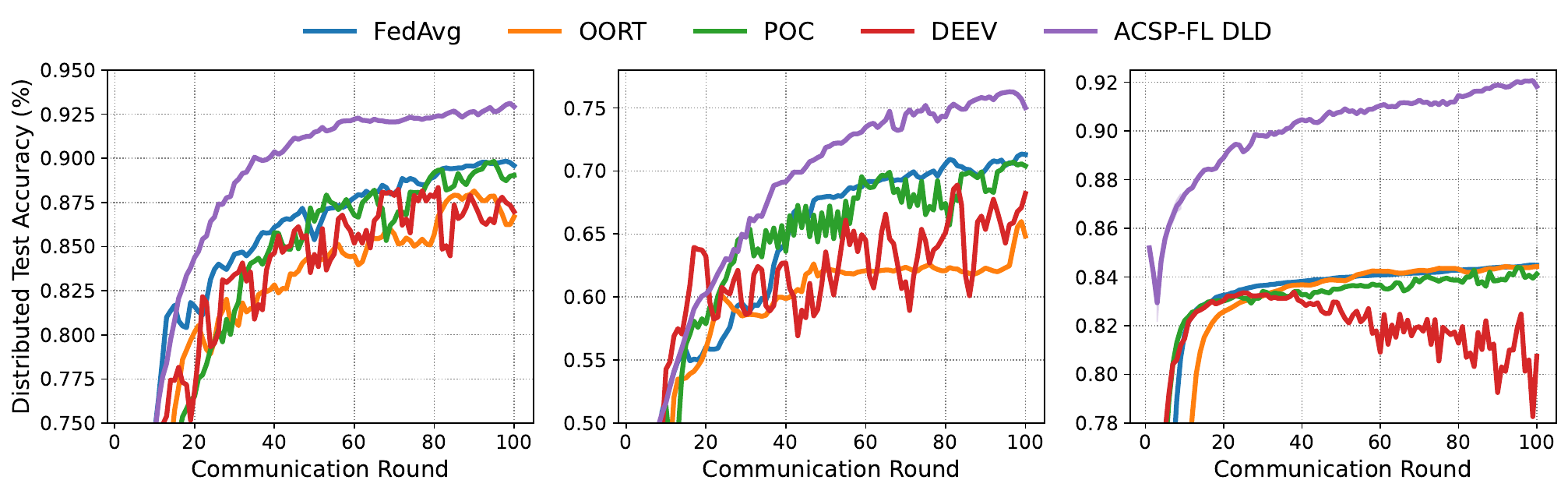}
    \caption{Test accuracy in function of the communication round for each dataset.}
    \label{fig:test_acc}
\end{figure*}

\textcolor{black}{Fistly, \Cref{fig:test_acc} shows distributed accuracy evaluation for the three datasets analyzed for each solution. An exploratory analysis was performed to define the best hyper-parameters for the literature solutions. Therefore, for the OORT and POC solution, which need to define the $k$ number of clients to be selected per round, for $k = 50\%$, we had the best cost-benefit between performance and communication overhead. On the other hand, we define the FedAvg as a baseline and thus define that 100\% of clients will always be selected in each round. Furthermore, for {\NAME}, we set the value for $decay = 0.005$.}

As seen in \Cref{fig:test_acc}, \NAME\ improved the model's performance for all datasets compared to the other solutions. This results from the model personalization used by \NAME\ that considers both local and global models to decide which is more suitable for each client. In addition, we can also see the advantages of layer sharing among clients to improve \NAME\ performance. \textcolor{black}{In particular, \NAME\ with both approaches increases the model's accuracy in up to $8\%$ when compared to FedAvg, POC, OORT, and DEEV. It is worth noticing that \NAME\ does not overfit the model for each client since when combining the personalization method with the layer sharing, a collaborative training is introduced, consequently \NAME\ improving its accuracy.}

\begin{figure*}[ht]
    \centering
    \includegraphics[width=\textwidth]{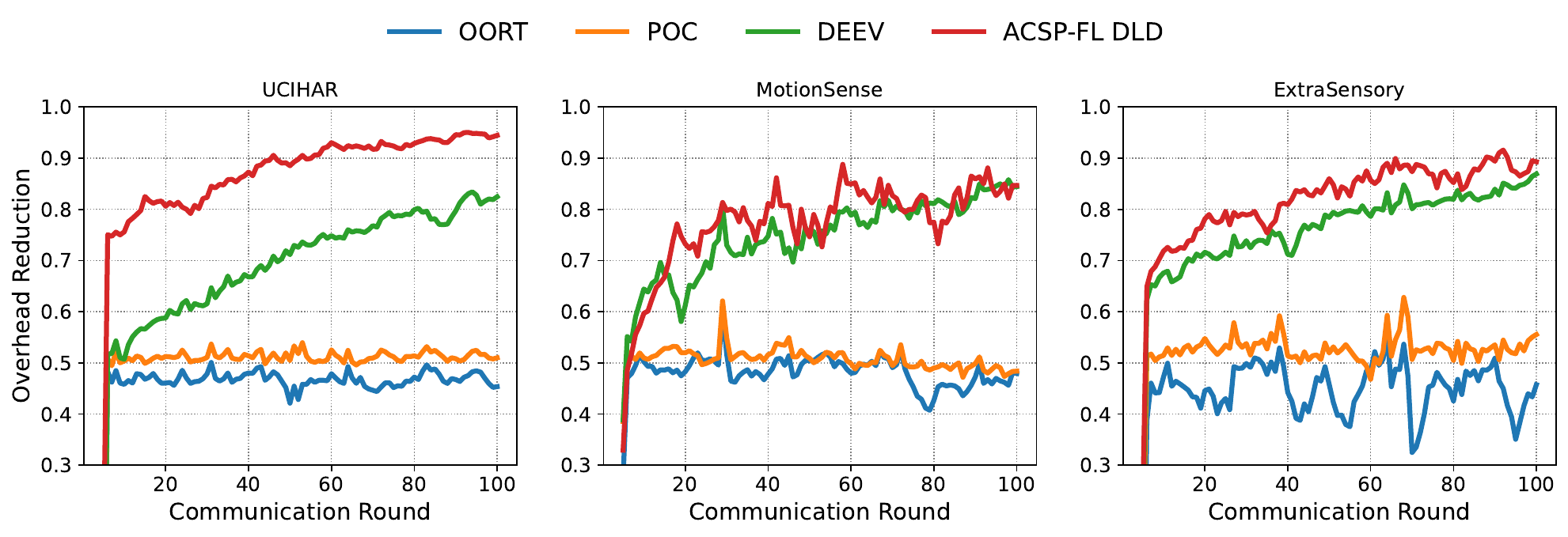}
    \caption{Latency produced by each solution for each dataset.}
    \label{fig:latency}
\end{figure*}
\begin{table*}[!ht]
\caption{Summary of assessed metrics for \NAME\ and literature solutions.}\label{tab:results_literature}
\resizebox{0.95\textwidth}{!}{%
\color{black}
\begin{tabular}{ccccccc}
\toprule
                                       &                    &                        & \multicolumn{4}{c}{\textbf{Metrics}}                                                                             \\ \midrule
\textbf{}                              & \textbf{Solutions} & \textbf{Accuracy (\%)} & \textbf{TX bytes (Mb)} & \textbf{TX bytes per client (Mb)} & \textbf{Convergence time (s)} & \textbf{Efficiency} \\ \midrule
\multirow{5}{*}{\textbf{UCI-HAR}}      & FedAvg             & 0.89                   & 992.52                 & 33.08                             & 19.17                         & 0.42                \\
                                       & OORT               & 0.87                   & 501.22                 & 16.70                             & 20.36                         & 0.70                \\
                                       & POC                & 0.88                   & 501.22                 & 16.70                             & 18.70                         & 0.69                \\
                                       & DEEV               & 0.87                   & 314.29                 & 10.47                             & 18.39                         & 0.69                \\
                                       & ACSP-FL DLD        & 0.92                   & 0.22                   & 0.007                             & 9.66                          & 0.95                \\ \midrule
\multirow{5}{*}{\textbf{MotionSense}}  & FedAvg             & 0.70                   & 113.26                 & 4.71                              & 240.82                        & 0.39                \\
                                       & OORT               & 0.65                   & 57.19                  & 2.38                              & 249.13                        & 0.62                \\
                                       & POC                & 0.70                   & 57.19                  & 2.38                              & 236.58                        & 0.61                \\
                                       & DEEV               & 0.67                   & 32.98                  & 1.37                              & 212.78                        & 0.73                \\
                                       & ACSP-FL DLD        & 0.75                   & 0.24                   & 0.010                             & 216.76                        & 0.87                \\ \midrule
\multirow{5}{*}{\textbf{ExtraSensory}} & FedAvg             & 0.84                   & 1114.17                & 18.56                             & 42.91                         & 0.44                \\
                                       & OORT               & 0.84                   & 562.65                 & 9.37                              & 47.31                         & 0.69                \\
                                       & POC                & 0.84                   & 562.65                 & 9.37                              & 40.38                         & 0.68                \\
                                       & DEEV               & 0.81                   & 311.41                 & 5.19                              & 35.73                         & 0.79                \\
                                       & ACSP-FL DLD        & 0.92                   & 0.28                   & 0.004                             & 38.91                         & 0.95     \\ \bottomrule          
\end{tabular}%
}
\normalcolor
\end{table*}

Another important metric to be assessed is the overhead produced by each solution, as depicted in \Cref{fig:latency}. The benefits of the dynamic client selection and layer sharing used by \NAME\ are evident, resulting in a reduction of approximately $90\%$ compared to FedAvg's overhead. In this study, overhead represents the time spent for each solution to receive a model from the server, perform a training round, and report it back to the server. Therefore, as fewer clients are selected according to the decay effect implemented by \NAME, the server must wait for fewer clients to proceed to the next communication round. Furthermore, as \NAME\ also shares fewer layers than the other solutions, it further reduces the communication overhead produced by model sharing, consequently lowering the overhead.

\textcolor{black}{Comparing \NAME\ with POC and OORT solutions that require determining the number of clients participating in each communication round in advance we observe that the overhead reduction is proportional to the number of clients selected by POC. This reduction remains almost constant since it does not change that number over the communication rounds. However, OORT achieves a lower reduction, as it needs to compute the statistical metric for each client in every single communication round. Lastly, when comparing \NAME\ with DEEV, a slight improvement results from layer sharing. Consequently, less time is spent transferring the model over the network. In particular, \NAME\ reduces latency by approximately $30\%$ and $10\%$ compared to POC and DEEV.}

\textcolor{black}{Table \ref{tab:results_literature} provides a comprehensive summary of the assessed metrics for \NAME\ compared to other solutions in the literature, addressing datasets such as UCI-HAR, MotionSense, and ExtraSensory. In terms of accuracy, \NAME\ DLD stands out, achieving 92\% in the UCI-HAR dataset, 75\% in MotionSense, and 92\% in ExtraSensory, surpassing other approaches. Additionally, when analyzing communication efficiency, \NAME\ DLD achieves a notable reduction in transmitted data and convergence time. Compared to FedAvg, it exhibits a reduction of approximately 99\% in transmitted bytes in UCI-HAR, 99\% in MotionSense, and 97\% in ExtraSensory. Furthermore, the \NAME\ DLD solution excels in efficiency, with a significant reduction in communication overhead, making it a promising option for performance optimization in federated learning environments.}

\textcolor{black}{Finally, the results show that \NAME\ DLD excels in striking a balance between communication overhead reduction and model accuracy. The significant efficiency gains, as indicated by metrics exceeding 0.9 in all cases, underscore the efficacy of the dynamic layer definition mechanism employed by \NAME\ DLD. By adapting the number of shared layers based on client accuracy, \NAME\ DLD optimizes communication efficiency, showcasing its robust performance in federated learning scenarios.}

\subsection{Clients Performance Analysis}
\begin{figure*}[ht]
    \centering
    \subfigure[UCI-HAR]{
        \label{hist:ucihar}
        \includegraphics[width=0.93\textwidth]{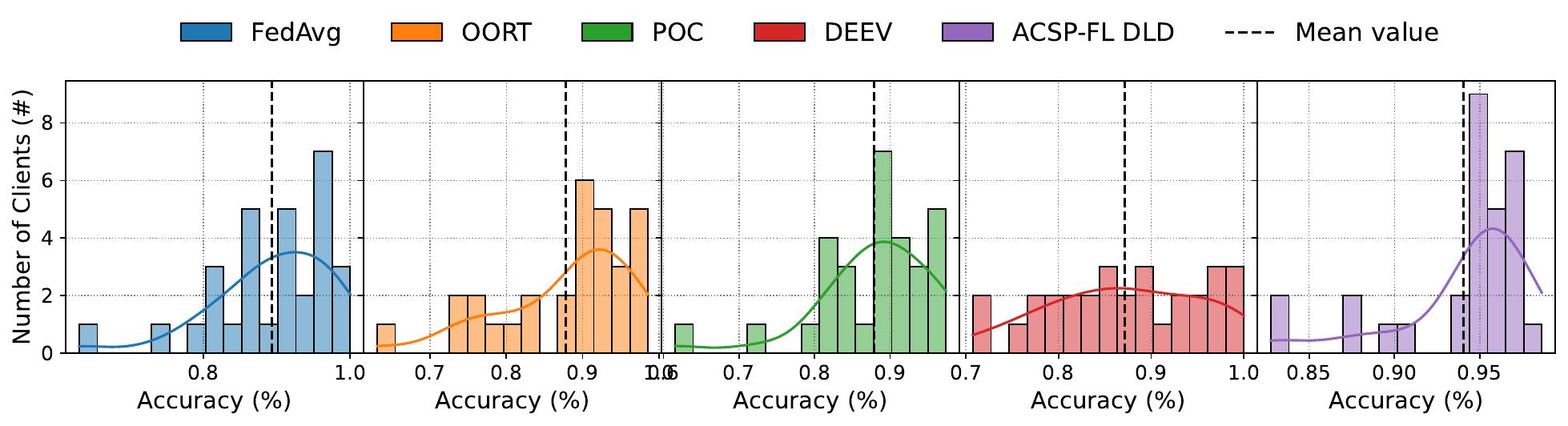}
    }
    \subfigure[MotionSense]{
        \label{hist:motionsense}
        \includegraphics[width=0.93\textwidth]{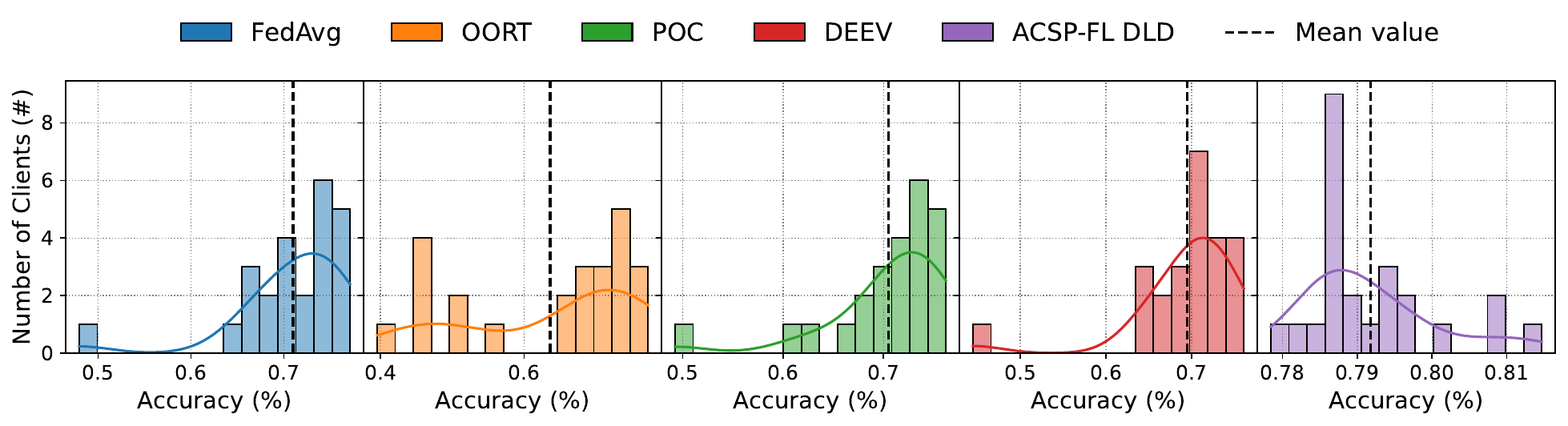}
    }
    \subfigure[ExtraSensory]{
        \label{hist:extrasensory}
        \includegraphics[width=0.93\textwidth]{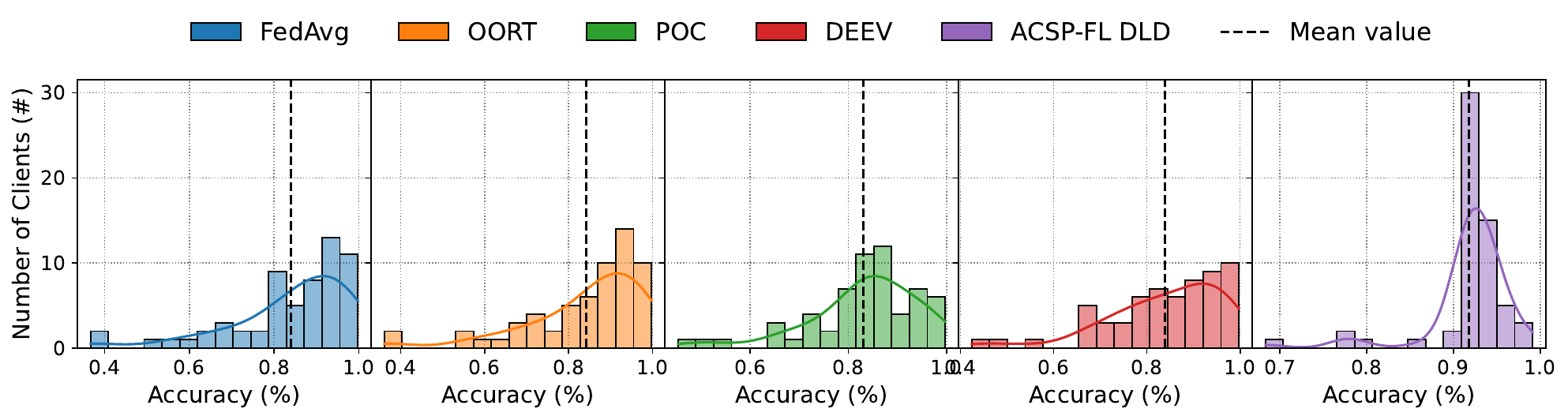}
    }
    \caption{Accuracy distribution for each client at the last communication round.}
    \label{fig:acc_clients}
\end{figure*}
\Cref{fig:acc_clients} illustrates the performance of each client for each dataset at the final communication round. The implemented personalization methods in \NAME\ are evident in the higher accuracy achieved for the majority of clients. This metric is crucial for depicting the overall model performance, as relying solely on the average accuracy of clients may hide the performance of specific clients with notably low accuracy. 

Therefore, when considering the number of clients higher than the average accuracy, we can see that for \NAME, more clients achieve accuracy higher than the average one, and the average value is higher than the average value of literature solutions. Moreover, the lowest accuracy for the clients is also higher in \NAME. For instance, when considering the \Cref{hist:extrasensory}, the client with lower accuracy in FedAvg is approximately $40\%$. However, when comparing to \NAME, the client with the lowest performance achieves accuracy higher than $70\%$. This data shows how efficient \NAME\ is in empowering clients to achieve better results with fine-tuning and layer sharing.

\subsection{Client Selection Analysis}

\begin{figure*}[]
    \centering
    \includegraphics[width=0.99\textwidth]{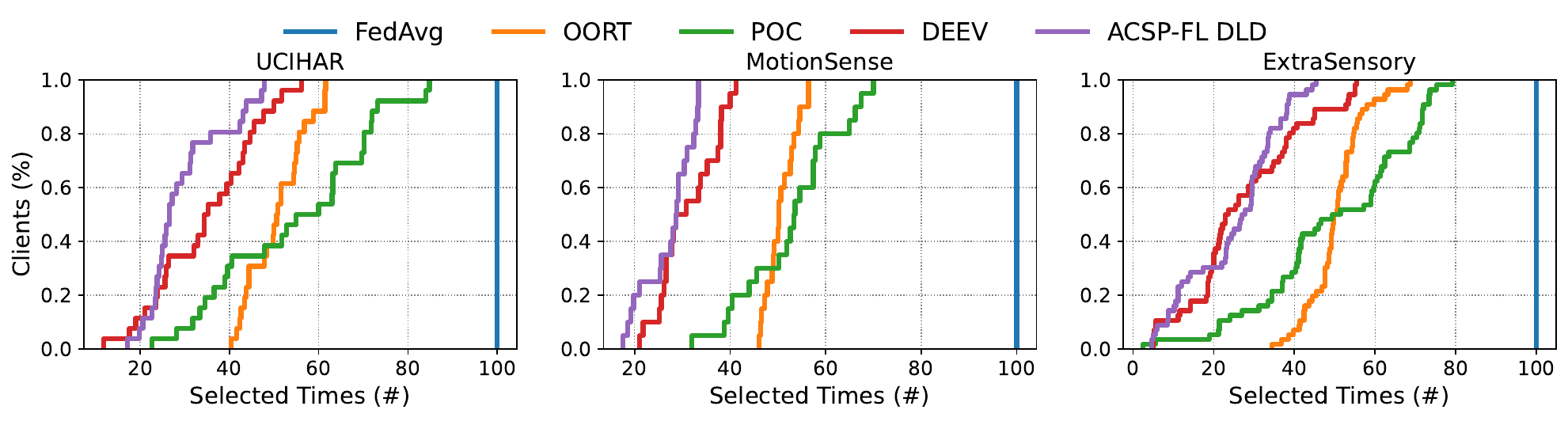}
    \caption{Evaluation of how many times clients were selected by each solution.}\label{fig:client_selection_results}
\end{figure*}

\textcolor{black}{\Cref{fig:client_selection_results} presents the client selection frequency for each solution. As discussed previously, FedAvg exhibits a 100\% selection rate for all clients across 100 rounds, a consequence of the simulation parameters that involve the participation of all clients in each communication round. In contrast, POC and OORT solutions specify a fixed number of clients for participation in each round, differing in their selection mechanisms. The results reveal that OORT selects approximately 40\% more clients compared to POC. However, for the other 50\% of clients, POC increases their participation frequency in the training phase. This outcome stems from POC's selection mechanism, exclusively utilizing the poorest-performing clients, leading to a consistent set of clients participating in subsequent rounds after the initial 50 rounds.}

\textcolor{black}{When examining \NAME\ with adaptive client selection, personalization, and layer sharing, clients are chosen less frequently for model training. This is a consequence of improved client performance, as depicted in \Cref{fig:acc_clients} where a greater number of clients exhibit performance surpassing the average. Consequently, fewer clients remain eligible for selection and model training, given that these solutions exclusively consider clients with performance lower than the average model performance. In comparison to DEEV, \NAME\ achieves a reduction of up to 30\% in the frequency of client selection. However, when juxtaposed with OORT, POC, and FedAvg, this reduction can reach up to 50\%, 60\%, and 70\%, respectively.}

\textcolor{black}{Upon examining the maximum number of selections, it is evident that within \NAME, clients were selected up to 50 times for UCI-HAR, approximately 35 times for MotionSense, and around 50 times for ExtraSensory. In contrast, other solutions reached up to 80 selections, as seen in the case of POC for ExtraSensory and UCI-HAR. It is noteworthy that DEEV achieves a similar performance when compared to \NAME\ for some datasets. It is consequence of its adaptive client selection mechanism. However, the slight improvement in \NAME\ is a result of its personalization method that enhances its performance.}

Finally, selecting fewer clients and fewer times improves the overall system performance because it will also reduce the communication and processing overheads. In addition, selecting clients fewer times consumes fewer resources for each client since the server does not frequently select them to train the model. Thus, when considering resource constraints scenarios, \NAME\ seems an excellent choice to preserve the clients' resources. 

With these results, we can conclude that: \textit{(i)}~\NAME\ provides an efficient adaptive client selection mechanism that reduces the overall latency of the system; \textit{(ii)}~the personalization employed by name enables clients to achieve better performance; \textit{(iii)}~personalization with layer sharing helps to improve model performance while reducing the communication overhead since fewer layers are transmitted over the network. 

\section{Conclusion}
\label{sec:conclusion}

This work presented \NAME, a federated learning solution with personalization and adaptive client selection based on the performance of the distributed model. \NAME~ also implements a gradual decay function to reduce the number of clients selected for each round according to model performance. This design leads to faster convergence and reduces processing and communication \textit{overhead}. Also, it implements a personalization method that enables clients to fine-tune the model for their dataset, and by using a layer sharing mechanism \NAME\ transmits fewer data (i.e., parameters of the model) over the network, consequently reducing the communication overhead even further.

To evaluate \NAME's performance, we used three different datasets for human activity recognition created with data from \textit{smartphones} sensors. The results showed that \NAME~reduces the communication overhead by up to 95\% and also the latency to train the federated model by more than 90\% concerning literature solutions. Furthermore, \NAME\ also increases the model's performance by at least $10\%$ with its personalization and layer-sharing approach compared to other literature solutions.

\textcolor{black}{Privacy is one key aspects in FL services, which not only aligns with ethical considerations but also fosters trust among participants, encouraging widespread participation in collaborative machine learning endeavors while preserving the confidentiality of individual data. In this way, it is important to stress that additional methods to improve even further clients' privacy  can be implemented in \NAME\, such as secure aggregation and differential privacy based algorithms.}

Finally, as future work, information about device capabilities will be incorporated to enable context-aware selection of battery, CPU and memory usage, and even network quality of devices. Model compression algorithms will also be developed to reduce the solution's communication overhead further.

\section*{Declaration of Competing Interest}
The authors declare that they have no known competing financial interests or personal relationships that could have appeared to influence the work reported in this paper.

\section*{Acknowledgments}
This project was supported by the Ministry of Science, Technology, and Innovation of Brazil, with resources granted by the Federal Law 8.248 of October 23, 1991, under the PPI-Softex. The project was coordinated by Softex and published as Intelligent agents for mobile platforms based on Cognitive Architecture technology [01245.013778/2020-21]. 

\section*{Data availability}
Data and code are available in the project repository\footnote{\url{https://github.com/AllanMSouza/ACSP-FL}}.

\printcredits

\bibliographystyle{model1-num-names}





\end{document}